\documentclass[letterpaper, 10 pt, conference]{ieeeconf}  

\IEEEoverridecommandlockouts                              

\usepackage{xcolor}
\usepackage{colortbl}

\usepackage{censor}

\usepackage{amsmath}
\usepackage{amssymb}

\usepackage{booktabs}
\usepackage{multirow}
\usepackage{subcaption}
\usepackage{threeparttable}

\usepackage{graphicx}
\usepackage[export]{adjustbox}

\usepackage{cite} 

\newcommand{\centerzoom}[3][0.333]{%
  \adjustbox{trim={{#1\width} {#1\height} {#1\width} {#1\height}},clip,width=#2}{\includegraphics{#3}}%
}

\usepackage{pifont}
\newcommand{\cmark}{\ding{51}}%
\newcommand{\xmark}{\ding{55}}%

\providecommand{\BackProj}{\mathcal{B}}

\providecommand{\PicWidthPage}{\def\svgwidth{\textwidth}}

\usepackage[inkscapelatex=false]{svg}
\graphicspath{{figures/}}

\usepackage[breaklinks,colorlinks,allcolors=black]{hyperref}

\title{\LARGE \bf
PIXIE: A Zero-Shot texture-invariant 6D pose estimation framework for unseen objects with assembly defects
}

\author{Leon Jungemeyer$^{1}$, Alejandro Magaña$^{2}$, Gautham Mohan$^{1}$, Matthias Karl$^{1}$, and Daniel Werdehausen$^{1}$%
\thanks{$^{1}$Leon Jungemeyer, Gautham Mohan, Matthias Karl, and Daniel Werdehausen are with Carl Zeiss AG, Oberkochen, Germany.}%
\thanks{$^{2}$Alejandro Magaña is with Carl Zeiss Digital Innovation, Munich, Germany.}%
\thanks{Corresponding authors: Leon Jungemeyer (leon.jungemeyer@zeiss.com) and Alejandro Magana (alejandro.magana@zeiss.com).}
}

\begin{document}

\maketitle

\begin{abstract}
6D pose estimation remains a key challenge in robotics and computer vision, particularly in industrial 
environments. The deployment of currently available data-driven methods is often limited by 
resource-intensive data pipelines, reliance on textured 3D models, and sensitivity to geometric 
deviations caused by damages or assembly defects.
We present PIXIE, a zero-shot framework that estimates the 6D pose of an object from an RGB image 
using only an untextured 3D model.
Synthetic depth and normal maps are rendered from sampled reference viewpoints and matched to the query 
image via a pretrained cross-modality feature matcher. Matched keypoints are back-projected to obtain 
2D--3D correspondences for PnP-based pose estimation.
Relying exclusively on geometry makes the method inherently robust to lighting and texture variation, 
while correspondence filtering handles geometric deviations between the model and physical object.
We evaluate on widely-used public benchmarks, reporting state-of-the-art results on texture-less 
objects without object-specific training, and introduce a novel dataset with assembly defects, 
texture variations, and occlusion to demonstrate real-world applicability.
\end{abstract}
\section{Introduction}
\label{sec:intro}
The detection of 3D objects and estimation of their 6D poses are fundamental tasks for automating various computer vision 
applications, ranging from robotic manipulation to industrial inspection.
Traditional and state-of-the-art methods typically rely on learning discriminative features from textured object 
appearances. These approaches require either extensive training data with annotations or textured 3D models that accurately 
represent the target objects.
However, in many practical scenarios, particularly in industrial environments where parts change frequently or 
exhibit high variety, textured 3D models are rarely available. 
Instead, only technical drawings or untextured 3D models are available, which capture geometric information but lack 
material and appearance details.
Additionally, these 3D models often deviate geometrically from the actual object instances due to manufacturing 
tolerances, assembly defects, or wear and tear.
These deviations further change the appearance of the objects, obscuring the already limited texture information.
This problem, combined with the need for solutions that avoid resource-intensive and time-consuming data 
pipelines—including data acquisition, annotation, training, and validation—presents a fundamental challenge 
to achieving accurate and efficient 6D pose estimation in dynamic industrial environments.

To address these challenges, we propose a zero-shot framework specifically designed for industrial texture-less settings,
where objects surface appearance may vary, untextured 3D models are available, and the models may deviate geometrically from the physical object.
Our key insight is to leverage geometric features exclusively, bypassing the texture domain entirely.
This is realized through a pipeline (detailed in Section~\ref{sec:method}) built around three design choices:
rendering only geometry-derived depth and normal maps rather than photorealistic images,
encoding these continuous geometric maps via colormaps for compatibility with pretrained vision backbones,
and iteratively selecting precomputed reference viewpoints to refine the pose estimate.
Together, these choices yield a system that is inherently invariant to changes in object texture, color, or surface finish,
and remains functional even when assembly defects introduce geometric deviations between the 3D model and the physical object.
To the best of our knowledge, no existing method combines training-free, geometry-only zero-shot 6D pose estimation from a single RGB image with robustness
to both texture variation and geometric deviations from the 3D model.
We demonstrate the effectiveness of our method through comprehensive evaluation on the BOP benchmark, where it
outperforms state-of-the-art methods on textureless objects. 
To demonstrate the practical applicability of our method in industrial inspection settings, 
we introduce a novel dataset designed to capture the complexities of real-world scenarios, including geometric and texture deviations, varying lighting conditions, and occlusions.

Our paper presents the following main contributions:
(1) A training-free, geometry-only 6D pose estimation framework for zero-shot pose estimation under the joint presence of arbitrary texture changes and typical levels of assembly-related geometric deviations from the 3D model,
as commonly encountered in real industrial environments.
(2) A geometric rendering and cross-modality matching pipeline that establishes robust correspondences between 
synthetic geometric representations and real RGB images, resilient to geometric deviations and occlusions.
(3) A novel 6D pose estimation dataset, designed to evaluate robustness to real-world challenges in industrial inspection settings, including 
model deviations, texture variations, lighting changes, and occlusions.
\section{State of the art}
\label{sec:sota}

6D pose estimation from RGB images is a fundamental problem in computer vision,
with applications in robotics, augmented reality, and industrial inspection.
Traditional approaches fall into two broad groups:
feature-based pipelines~\cite{lepetit2009epnp, collet2011moped} that detect and match local descriptors to establish 2D--3D correspondences,
and learning-based methods~\cite{xiang2018posecnn, tremblay2018deep, zakharov2019dpod, magana2020posenetwork, nguyen2021fast}
that use deep neural networks to directly regress object pose or predict intermediate representations.
These families typically require substantial annotated data and struggle on texture-less or weakly textured industrial parts lacking distinctive appearance cues.

Render-and-compare and generalizable frameworks reduce object-specific supervision
by synthesizing viewpoints or leveraging few-shot priors.
Representative examples include OnePose / OnePose++~\cite{sun2022onepose, he2022onepose}, Gen6D~\cite{liu2022gen6d}, MegaPose~\cite{labbe2022megapose}, GigaPose~\cite{nguyen2024gigapose}, GenFlow~\cite{moon2024genflow}, and refinement pipelines like Pos3R~\cite{deng2025pos3r}.
While these improve generalization to novel instances,
they still depend on learned appearance priors, keypoint abstractions, or large render banks.
Benchmark efforts such as BOP~\cite{hodan2018bop} highlight persistent failure modes on low-texture and specular objects.

Zero-shot 6D pose estimation pushes further: estimating pose of unseen objects without object-specific training.
Foundation and universal correspondence systems (ZS6D~\cite{ausserlechner2024zs6d}, FoundPose~\cite{ornek2024foundpose}, Pos3R~\cite{deng2025pos3r})
leverage broadly pre-trained visual representations or matching transformers.
They increase robustness to shape variation and partial occlusion,
but remain appearance dependent and thus sensitive to texture and color changes, lighting variation, and material reflectance.

Detector-free architectures (SuperGlue~\cite{sarlin2020superglue}, LoFTR~\cite{sun2021loftr}) illustrate the power of learned correspondence priors.
To mitigate appearance dependence, cross-modality matching has emerged~\cite{ren2025minima, li2024matching, he2025matchanything}.
In particular, Ren et al.~\cite{ren2025minima} train a modality-invariant matcher to align real RGB images with synthetic representations of the scene, including depth and normal maps.
They use a synthetic data engine that generates training pairs from public sources (e.g., MegaDepth~\cite{li2018megadepth}) and show transfer to diverse downstream tasks.
However, these methods focus only on generating 2D--2D correspondences and do not address the specific challenges of 6D pose estimation, 
such as handling geometric deviations between the 3D model and the physical object, or achieving robustness to occlusion and texture variation in industrial settings.

Current literature lacks a \emph{training-free}, geometry-only zero-shot pipeline
that (i) operates on 2D images and untextured 3D models,
(ii) uses only 3D renderings without model-specific training, 
(iii) is inherently invariant to texture and lighting, 
and (iv) tolerates moderate geometric deviations. 

\section{6D Object Pose Estimation Using Cross-Modality Feature Matching}
\label{sec:method}

Our method provides a \emph{training-free} framework for zero-shot 6D pose estimation of 3D objects with unknown textures as well as geometric deviations---a common setting in industrial scenarios.
It relies \textit{exclusively} on geometric renderings (depth and surface normal maps) of the 3D model and cross-modality feature matching, eliminating appearance-based domain gaps and the need for domain randomization, photorealistic rendering, or any instance-specific training.

We detail reference generation (§\ref{sec:method_reference_generation}), cross-modality matching (§\ref{sec:method_feature_matching}), pose estimation (§\ref{sec:method_pose_estimation}), and an iterative refinement (§\ref{sec:method_sampling_poses}).

We refer to our method as \textbf{PIXIE}: \textbf{P}ose \textbf{I}dentification e\textbf{X}cluding appearance with \textbf{I}mperfect g\textbf{E}ometry.

\subsection{Task Definition and Constraints}
\label{sec:method_task}
Given a query image $I_q$ containing an object and its corresponding 3D model $M$, our goal is to estimate the 
6D object pose $T^{\text{O}} \in SE(3)$—comprising rotation $R \in SO(3)$ and translation $t \in \mathbb{R}^3$—relative to the camera 
frame. Let $K$ denote the intrinsic matrix of a pinhole camera model, containing focal lengths and principal point.

\textbf{Key capabilities and design constraints:}
Our framework is designed to satisfy the following properties:
\begin{enumerate}
    \item \textbf{Zero-shot operation:} No training is required on object $M$ or any of its instances. 
    The cross-modality feature matcher is pre-trained on large-scale generic datasets but has not seen the 
    specific object during training.
    
    \item \textbf{Texture invariance:} We rely exclusively on geometry-derived representations (depth and normal 
    maps) from the 3D model. Changes in object appearance, such as different surface finishes, paint colors, 
    lighting conditions, wear and tear, or geometric deviations, do not require additional reference generation or model updates.
    
    \item \textbf{Robustness to geometric variations:} Small deviations between the 3D model and the physical 
    object (e.g., manufacturing tolerances, minor wear, or assembly imperfections) are handled through robust 
    filtering.     
    
\end{enumerate}

\subsection{Zero-Shot Pose Estimation Pipeline}
\label{sec:zero_shot_methodology}
The approach operates in two stages: an offline reference generation phase and an online inference phase.
During inference, the pipeline processes the query image through several steps to establish correspondences, 
estimate the pose, and iteratively refine the result.

Concretely: (1)~geometric reference views (depth and normal maps) are rendered offline from canonical viewpoints around $M$; (2)~optionally, a detector crops $I_q$ to the object region; (3)~a cross-modality matcher establishes 2D--2D correspondences between $I_q$ and each reference view; (4)~correspondences are lifted to 2D--3D and a PnP+RANSAC solver yields an initial pose; and (5)~nearby precomputed views are iteratively selected and re-matched to refine the estimate.

\begin{figure*}[t]
    \centering
    \PicWidthPage
    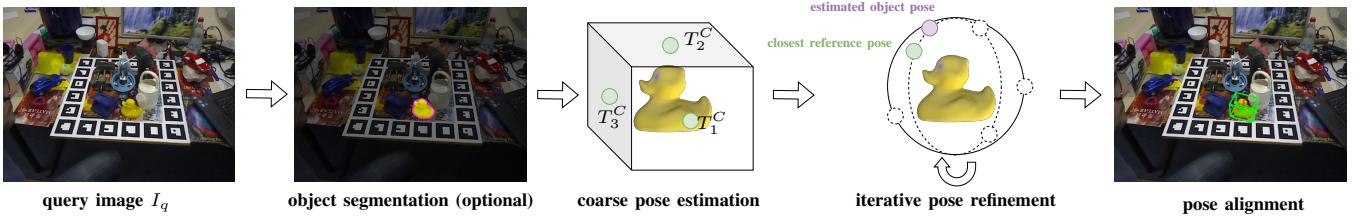
    \caption{
        Inference pipeline overview. 
        Optional object localization extracts the region of interest from the query image.
        Pre-generated geometric reference views (depth and normal maps) are matched against the query using a 
        cross-modality feature matcher.
        Matched keypoints are lifted to 3D using known rendering geometry, enabling PnP-based pose estimation with 
        RANSAC.
        Iterative view selection further improves accuracy by re-matching against precomputed locally sampled views.
    }
    \label{fig:inference_overview}
\end{figure*}

Figure~\ref{fig:inference_overview} illustrates the inference pipeline from query image to final pose estimate.

\subsection{Geometric Reference Image Generation}
\label{sec:method_reference_generation}
To enable cross-modality matching between RGB query images and the 3D model, we generate a set of geometric reference views 
offline.
These references encode only the object's shape and surface properties, avoiding any dependence on texture or 
material appearance.
For each view, they include \emph{depth} and \emph{surface normal} maps, 
providing complementary geometric information: depth captures the overall 3D structure, while normals encode 
fine-scale surface details. 

\textbf{Rendering geometric maps:}
For each reference viewpoint $i$, let $T^C_i = [R_i | t_i]$ denote the object-to-camera transformation and 
$K^{start}$ denote the camera intrinsic matrix used for rendering.
Note that $K^{start}$ does not need to match the actual query camera intrinsics $K$; 
choosing $K^{start}$ reasonably close to $K$ improves matching, but exact equality is not required.

The depth map $D_i$ and normal map $N_i$ are rendered from the selected viewpoints $i = 1, \ldots, N$, 
projecting the model $M$ at pose $T_i$ onto the image plane.

\begin{figure}[t]
    \def\svgwidth{\columnwidth}
    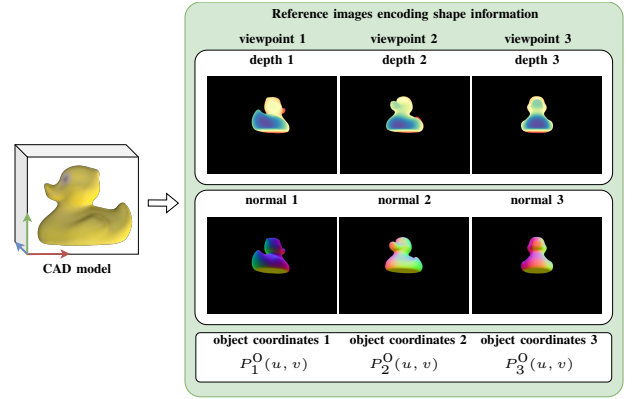
    \caption{
        Examples of geometric reference views generated from the 3D model.
        In the top row, depth maps capture overall 3D structure; in the bottom row, normal maps encode surface orientations.
        Both representations are normalized and colormapped for compatibility with cross-modality feature matchers. 
    }
    \label{fig:reference_overview}
\end{figure}

\textbf{Encoding for cross-modality matching:}
A pre-trained cross-modality feature matcher such as MINIMA or MatchAnything~\cite{ren2025minima, he2025matchanything} is used to match geometric references to RGB images. 
These matchers typically expect image inputs in 8-bit RGB format.
Thus, depth and normal values are normalized to an 8-bit range and a fixed colormap 
is applied.
This preserves the relative geometric structure while enabling the use of generic vision backbones trained on 
natural images.
Figure~\ref{fig:reference_overview} shows examples of the rendered depth and normal maps.

\textbf{Back-projection and 3D correspondence lookup:}
Since the rendering process is fully controlled, each valid pixel in the depth map can be back-projected to 
obtain its 3D coordinates.
For a pixel at $(u,v)$ with depth $D_i(u,v)$, the camera-space 3D point is first computed as:
\begin{equation}
    \mathbf{P}_i^{\text{cam}}(u,v) = \BackProj\big(u, v, D_i(u,v), K_{start}\big),
\end{equation}
where $\BackProj$ applies the inverse camera projection using intrinsics $K_{start}$.
This is then transformed to object coordinates:
\begin{equation}
    \mathbf{P}_i^{\text{O}}(u,v) = (T^C_i)^{-1} \mathbf{P}_i^{\text{cam}}(u,v)
\end{equation}
This back-projection enables lookup of the 3D object-space point corresponding to any matched keypoint in the 
reference view.

\textbf{Reference view data structure:}
For each reference viewpoint $i$, the following data is stored:
\begin{equation}
    \mathcal{R}_i = \left\{D_i, N_i, \mathbf{P}_i^{\text{O}}, T^C_i, K^{start}\right\},
\end{equation}
where $D_i$ is the rendered depth map, $N_i$ is the normal map, $\mathbf{P}_i^{\text{O}}$ contains the 
back-projected 3D points for all pixels, $T^C_i$ is the rendering pose, and $K^{start}$ are the rendering intrinsics.

\textbf{Viewpoint sampling strategy:}
Reference views are generated from multiple viewpoints around the object to ensure sufficient coverage.
Six coarse views are rendered from canonical orientations aligned with the object's bounding box, providing 
broad initial coverage.
Additional fine-grained viewpoints are uniformly sampled on a sphere surrounding the object using a Fibonacci lattice~\cite{gonzalez2010measurement}.

\subsection{Cross-Modality Feature Matching}
\label{sec:method_feature_matching}

Given the geometric reference views, correspondences are established between 
the query image and the synthetic geometric renderings.
This matching problem is addressed using recent cross-modality neural feature matchers trained on diverse image 
modalities~\cite{ren2025minima, he2025matchanything}.
These models are trained on a wide variety of image types, including RGB images, depth maps, normal maps, and 
infrared images, enabling them to find correspondences across modalities with little visual similarity.
Our framework is agnostic to the specific cross-modality matcher.

\textbf{Matching process:}
For each reference view $\mathcal{R}_i$, the matcher processes the query image $I_q$ together with the encoded 
depth map $D_i$ and normal map $N_i$.
The matcher outputs a set of keypoint correspondences $\{(u_q^j, v_q^j) \leftrightarrow (u_i^j, v_i^j)\}_{j=1}^{M_i}$ 
between the query image and reference view $i$, along with confidence scores for each match.

\textbf{Handling in-plane rotation:}
Cross-modality feature matchers can be sensitive to large in-plane rotations between the query and reference images.
To address this, each reference view is rotated around the viewing axis in discrete increments, and matching is performed for each rotated version.
This ensures that correspondences can be found even when the object appears rotated in the query image relative 
to the reference viewpoint.

\begin{figure}[b]
    \def\svgwidth{\columnwidth}
    \centering
    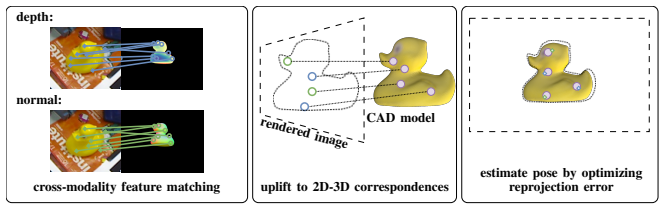
    \caption{
        Cross-modality feature matching establishes correspondences between the query RGB image and geometric 
        reference depth maps.
        These correspondences are uplifted to 2D-3D matches to enable 6D pose estimation via the PnP algorithm.
    }
    \label{fig:pose_estimation_overview}
\end{figure}

\subsection{Pose Estimation}
\label{sec:method_pose_estimation}

Given the 2D--2D keypoint correspondences from cross-modality matching, the object pose is estimated by lifting 
these to 2D--3D correspondences and applying a PnP (Perspective-n-Point) solver.
This approach builds upon established PnP and RANSAC-based methods~\cite{lepetit2009epnp, collet2011moped} and 
recent advances in robust pose estimation~\cite{deng2025pos3r}.

\textbf{Lifting 2D--2D to 2D--3D correspondences:}
For each matched keypoint pair $(u_q^j, v_q^j) \leftrightarrow (u_i^j, v_i^j)$ between the query image and 
reference view $i$, the reference keypoint is back-projected to 3D object coordinates using the stored data 
$\mathbf{P}_i^{\text{O}}$ from $\mathcal{R}_i$:
\begin{equation}
    \mathbf{X}_j = \mathbf{P}_i^{\text{O}}(u_i^j, v_i^j),
\end{equation}
yielding a 2D-3D correspondence $(u_q^j, v_q^j) \longleftrightarrow \mathbf{X}_j$ in object coordinates.
This process converts each 2D match into a correspondence between a query image pixel and a known 3D point on 
the object model. Figure~\ref{fig:pose_estimation_overview} illustrates this lifting process.

\subsection{Iterative View Selection with Precomputed Poses}

\label{sec:method_sampling_poses}

While the PnP solver from the previous subsection produces an initial pose estimate $T^{\text{O}}_0$ using the first 
set of six coarse reference viewpoints, this estimate may be insufficiently accurate.
When generating $T^{\text{O}}_0$, correspondences from all viewpoints can be concatenated into a single PnP problem, 
or separate pose estimates can be computed per viewpoint with the best-matching one selected based on inlier count.
The latter approach often performs better for objects with symmetries or ambiguous geometries.

To improve the initial estimate, an iterative view selection strategy is employed, selecting from precomputed candidate poses.
After obtaining $T^{\text{O}}_0$, the reference view $T^C_i$ whose camera optical axis best aligns with the optical axis of 
the estimated pose is selected.
Closeness is defined by the angular distance between the camera viewing directions of the two poses, i.e., the 
view with minimum angular deviation from $T^{\text{O}}_0$.

From the new reference view, the matching and pose estimation steps are repeated to refine the pose estimate.
This process generates a new pose $T^{\text{O}}_1$, which can be used as the basis for the next iteration.

Each selection iteration produces an updated pose $T^{\text{O}}_{(k+1)}$ by re-running feature matching and pose estimation 
with the selected reference views.
The process terminates when either (i) the selected best-matching view converges across iterations, or (ii) a 
maximum number of iterations is reached.
All viewpoints are precomputed offline, so no additional rendering or generation occurs during inference.
\section{Evaluation}
\label{sec:evaluation}

In this section, we present the evaluation of the proposed zero-shot, geometry-only pose estimation framework through two complementary 
experimental investigations.
In subsection~\ref{subsec:public_dataset}, we present quantitative results on selected datasets from the BOP benchmark~\cite{hodan2018bop}, 
comparing performance against state-of-the-art RGB-based 6D pose estimation methods, with particular focus on 
textureless objects where appearance-based approaches struggle.
In subsection~\ref{subsec:custom_dataset}, we demonstrate practical applicability through a custom dataset designed to capture real-world 
challenges in industrial inspection settings, including geometric deviations, texture variations, lighting changes, and occlusions.
This dataset significantly extends existing public datasets by systematically varying both texture and geometry.

\subsection{Experimental Setup}
\label{subsec:experimental_setup}

\textbf{Public benchmark configuration:}
Our experimental setup for public datasets follows standard protocols from related work on RGB-based 6D pose 
estimation, using identical datasets~\cite{hodan2018bop} and evaluation metrics to ensure objective comparison.
For each object model, we generate a single set of geometric reference views using the first camera calibration entry in each dataset.
Since camera calibrations vary across query images in these datasets, the reference and query images typically 
have slightly different intrinsics.
As demonstrated in Subsection~\ref{subsec:public_dataset}, our method can handle this mismatch robustly.

We used the CNOS~\cite{nguyen2023cnos} detector to obtain segmentation masks for all reference and query images, 
which are used to filter keypoints during feature matching and pose estimation.

\textbf{Custom dataset configuration:}
Our custom dataset is specifically designed to stress test the method under conditions that mimic industrial inspection scenarios.
In these scenarios, objects may exhibit varying textures, assembly defects, and other geometric deviations.

We used bounding-boxes provided with the dataset to crop the reference and query images.
These bounding-boxes are not perfectly tight and may include background clutter, 
which further tests the method's robustness to real-world conditions.

\textbf{Reference view generation:}
For all experiments, we generated 6 coarse reference views from canonical orientations (front, back, left, 
right, top, bottom) aligned with the object's bounding box.
Additionally, we sampled 115 fine reference views uniformly on a sphere surrounding the object using Fibonacci 
lattice sampling~\cite{gonzalez2010measurement}.
The sphere radius was determined based on the object's bounding box to ensure proper framing in rendered views.
All reference images were generated offline from the 3D model prior to evaluation, requiring no rendering at inference time.

To handle in-plane rotations, each reference view was augmented with 8 rotated versions at 45-degree increments, 
as described in Subsection~\ref{sec:method_feature_matching}.

\textbf{Feature matching and pose estimation:}
For cross-modality feature matching, we employed MINIMA~\cite{ren2025minima} with the LightGlue backbone~\cite{lindenberger2023lightglue}, 
a pre-trained matcher supporting diverse image modalities.

For pose estimation, we followed the RANSAC-based PnP solver detailed in Section~\ref{sec:method}, configured with 
a maximum of 1000 iterations, a reprojection error threshold of 5 pixels, and a minimum inlier ratio of 0.5.

\textbf{View selection:}
In all our experiments, we used only a single iteration of view selection, which already provided significant improvement over the initial estimate.
All viewpoints were precomputed offline, so no additional rendering or generation occurs during inference.

\subsection{Results on Public Datasets}
\label{subsec:public_dataset}

\begin{table*}[t]
\vspace{4pt}
\centering
\begin{threeparttable}
\begin{tabular}{l l c c|c c c c|c|c}
    \toprule
    & & & & \multicolumn{5}{c|}{Dataset (\# test images)} & \\
    \textbf{\#} & \textbf{Method} & \textbf{Training Free} & \textbf{Geometry Only} & \textbf{LM-O} & \textbf{T-LESS} & \textbf{TUD-L} & \textbf{ITODD} & \textbf{Custom Dataset} & \textbf{time (s)} \\
    & & & & 1145 & 6423 & 600 & 3041 & 695 &  \\
\midrule
1 & GigaPose \cite{nguyen2024gigapose} & {\color{red}\xmark} & {\color{red}\xmark} &    29.9 & 27.3 & 30.2  & 18.8 & 11.0\tnote{2}  & 0.9 \\
2 & GenFlow \cite{moon2024genflow} & {\color{red}\xmark} & {\color{red}\xmark} &       25.0 & 21.5 & 30.0 & 15.4 & - & 3.8\\
3 & MegaPose \cite{labbe2022megapose} & {\color{red}\xmark} & {\color{red}\xmark} &    22.9 & 17.7 & 25.8 & 10.8 & - & 15.5\\
4 & ZS6D \cite{ausserlechner2024zs6d}        & {\color{green}\cmark} & {\color{red}\xmark} &  29.8 & 21.0 & - & - & - & - \\
5 & FoundPose \cite{ornek2024foundpose} & {\color{green}\cmark} & {\color{red}\xmark} &    {\cellcolor{green!20}\textbf{39.6}} & 33.8 & {\cellcolor{green!20}\textbf{46.7}} & 20.5  & 53.2\tnote{2}  & 1.7 \\
6 & Pos3R \cite{pose3rcnn} & {\color{green}\cmark} & {\color{red}\xmark} &    30.5 & 23.6 & 43.2 & 25.1 & - & 1.4 \\
\midrule
\textbf{7} & \textbf{PIXIE (ours)} & \textbf{{\color{green}\cmark}} & \textbf{{\color{green}\cmark}}  & 19.4 & \cellcolor{green!20}\textbf{43.7} & 35.1 & \cellcolor{green!20}\textbf{26.7} & \cellcolor{green!20}\textbf{78.4} & 3.6\tnote{1} \\
\textit{8} & \textit{PIXIE - depth only (ours)} & \textit{{\color{green}\cmark}} & \textit{{\color{green}\cmark}}  & \textit{12.5} & \textit{38.2}  & \textit{28.5} & \textit{25.3} & \textit{71.8} & \textit{2.1\tnote{1}} \\
\textit{9} & \textit{PIXIE - normal only (ours)} & \textit{{\color{green}\cmark}} & \textit{{\color{green}\cmark}}  & \textit{17.6} & \textit{39.9} &\textit{28.6} & \textit{22.4} & \textit{71.1} & \textit{2.2\tnote{1}} \\
\bottomrule
\end{tabular}
\begin{tablenotes}\footnotesize
    \item[1] Measured on an NVIDIA RTX 4090 GPU with an AMD Ryzen 9 7950X CPU and 64\,GB of RAM.
    \item[2]{Using the official implementation with default settings.}
\end{tablenotes}
\caption{6D pose estimation comparison on BOP datasets (Average Recall, AR~\%~\cite{bop_benchmark}), benchmark numbers and runtimes obtained from Pos3R \cite{deng2025pos3r} and custom dataset results from our experiments.
Segmentation masks from CNOS~\cite{nguyen2023cnos} were used for all BOP methods (including our method), bounding boxes for the custom dataset.
To ensure a fair comparison, no refinement step was applied for any method, as refinement is typically heavily reliant on object textures.
Rows 8--9 are single-modality ablations (depth-only, normal-only).
}
\label{table:public_datasets_results}
\end{threeparttable}
\end{table*}

Methods requiring RGB-D input or non-zero-shot supervision are outside our scope because they use additional information unavailable to our setting.

\textbf{Performance on textureless objects:}
Our method demonstrated particularly strong performance on textureless datasets where geometric cues dominate.
On T-LESS, our geometry-only approach achieved 43.7\% Average Recall, outperforming all training-free methods 
and surpassing several appearance-based approaches that require object-specific training.
Similarly, on ITODD, our method achieved 26.7\% AR, exceeding all compared methods including those with 
training requirements.
On TUD-L, our method reached 35.1\% AR, demonstrating competitive results on texture-limited industrial objects.
Both modalities (rows 8--9) contribute: depth captures global 3D structure, while normals add complementary local surface detail.

\textbf{Performance factors on textured datasets:}
Performance on LineMOD Occluded (19.4\% AR) was lower compared to appearance-based methods, which is expected given that 
this dataset contains objects with distinctive textures that provide strong discriminative cues.
It is important to note that these methods leverage texture information, which is absent in our geometry-only approach, leading to a fundamental performance gap on this dataset.
Additionally, LM-O images have relatively low resolution (640\(\times\)480), which in combination with heavy occlusion and small objects limits the number and quality of geometric correspondences that can be established.

\textbf{Comparison without refinement:}
Many methods use a refinement step, often RGB-D or texture-based.
For fair zero-shot geometry-only comparison, we did not apply refinement in any of these evaluations (including our method).

\begin{figure}[htbp]
    \centering
    \includegraphics[width=0.22\textwidth]{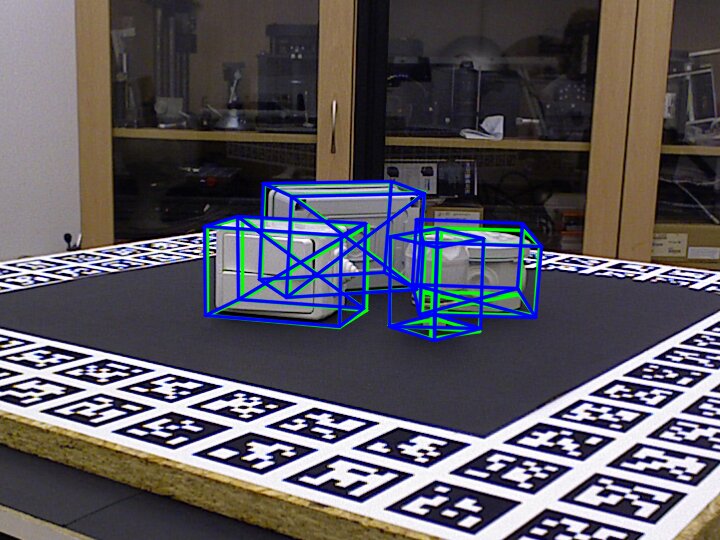}
    \includegraphics[width=0.22\textwidth]{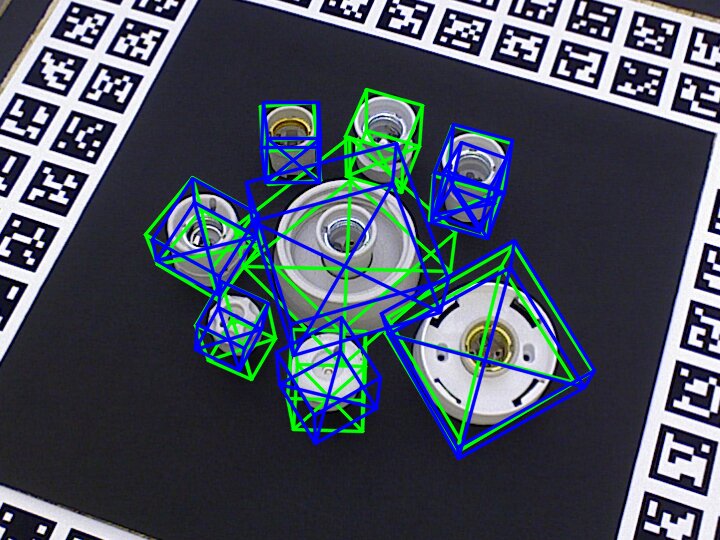}
    \includegraphics[width=0.22\textwidth]{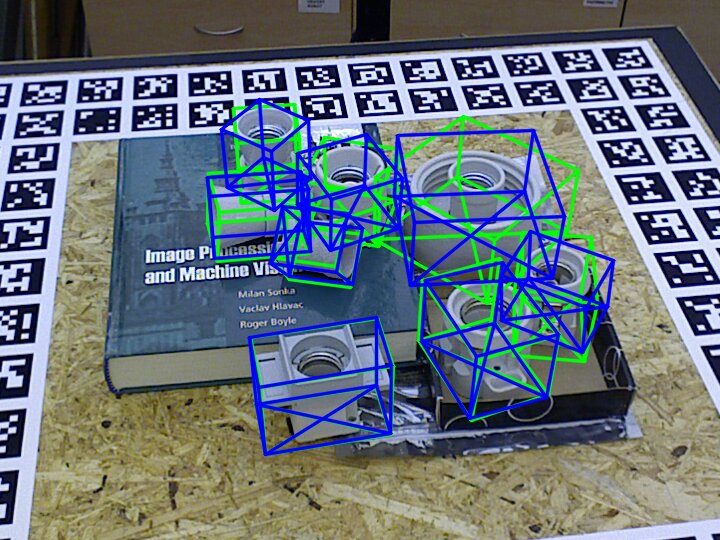}
    \includegraphics[width=0.22\textwidth]{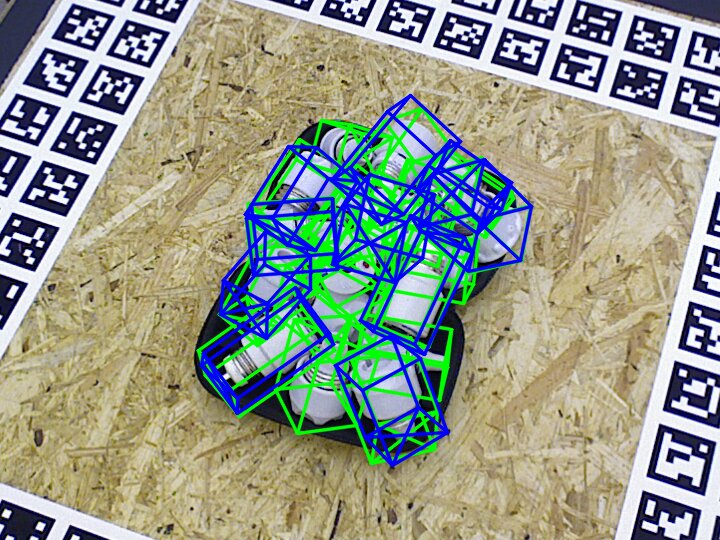}

    \caption{
        Qualitative results of our method on the T-LESS dataset from the BOP benchmark.
        Bounding boxes indicate ground truth poses (\textit{green}) and estimated poses (\textit{blue}).
        The geometry-only approach successfully estimates poses under significant occlusion and cluttered backgrounds 
        (left three examples).
        In highly cluttered scenes with severe occlusion, insufficient reliable correspondences can limit pose 
        accuracy (bottom right). 
    }
    \label{fig:bop_results}
\end{figure}

Figure~\ref{fig:bop_results} presents qualitative results of our method on the T-LESS dataset, illustrating successful pose 
estimation under occlusion and clutter, as well as failure cases where limited correspondences prevent accurate 
estimation.

\subsection{Results on Custom Datasets}
\label{subsec:custom_dataset}

Since public datasets do not capture the specific variations in texture and geometry required for our evaluation, we created a custom dataset of 695 images across 48 scenes capturing 4 target objects 
each with 3 distinct colored instances and up to 7 defective assemblies (Table~\ref{tab:dataset_summary}).
The dataset is publicly available at \href{https://zeiss.github.io/PIXIE-dataset/}{zeiss.github.io/PIXIE-dataset/}.
A target object is defined as an assembly of correctly mounted components with a corresponding 3D \emph{target model}, 
while the corresponding 3D models of defective assemblies are referred to as \emph{defective models}.

\begin{figure}[htbp]
    \centering
    \vspace{4pt}

    \begin{subfigure}{\linewidth}
        \centering
    
        \includegraphics[width=0.15\textwidth]{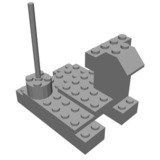}
        \includegraphics[width=0.15\textwidth]{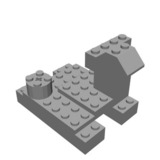}
        \includegraphics[width=0.15\textwidth]{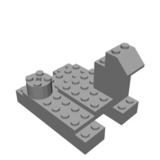}
        \includegraphics[width=0.15\textwidth]{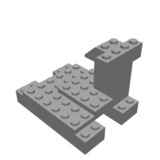}
        \includegraphics[width=0.15\textwidth]{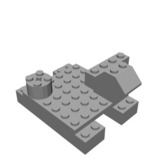}
        
        \caption{Different states of the same 3D model, including the \textit{target model} (left) and variations with \textit{assembly defects} (right four), which include missing components, misaligned parts, and incorrect assembly.
        }
        \label{fig:custom_industrial}
    \end{subfigure}\\[0.9em]

    \begin{subfigure}{\linewidth}
        \centering
        \centerzoom[0.25]{0.30\textwidth}{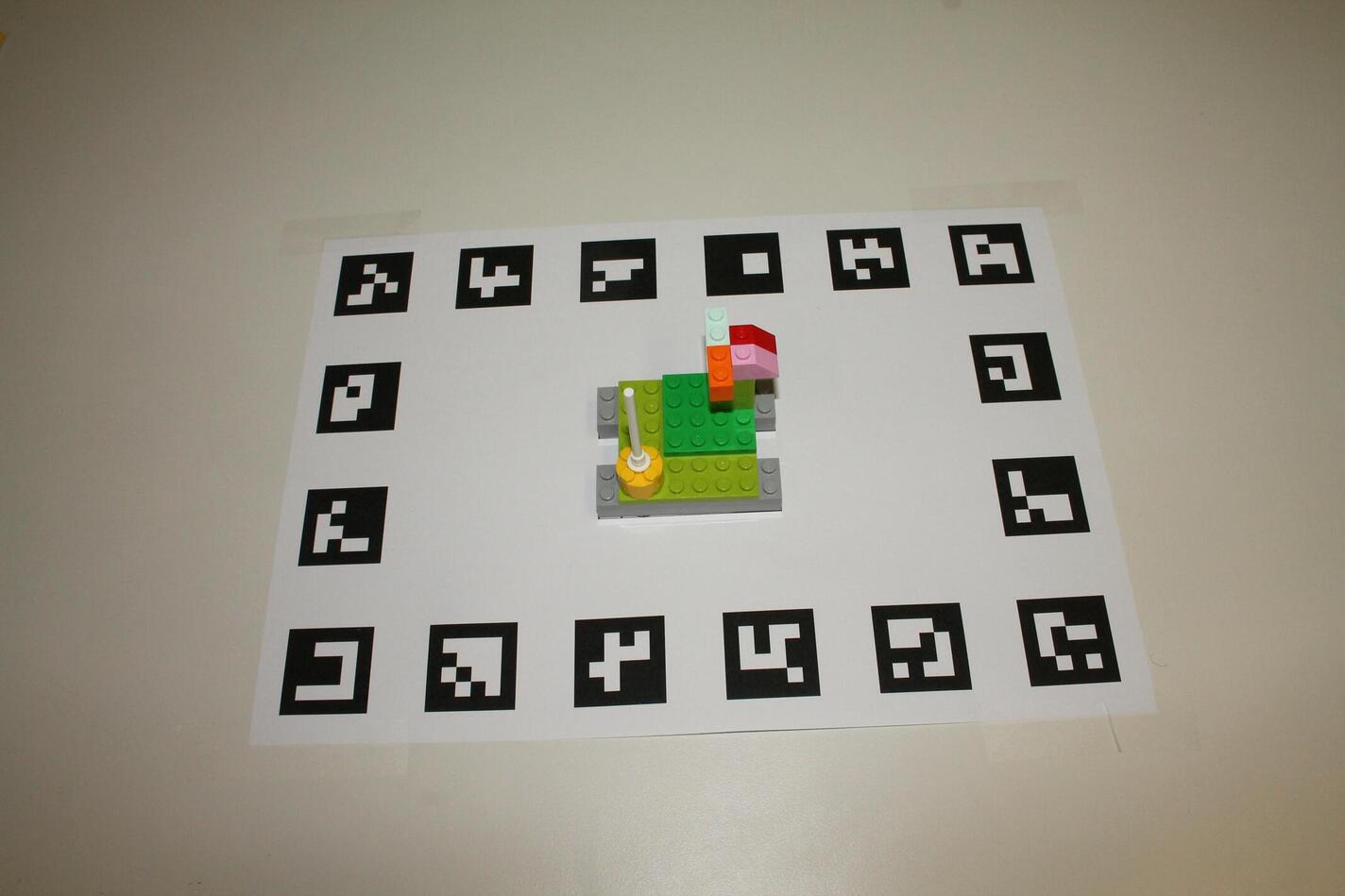}
        \centerzoom[0.25]{0.30\textwidth}{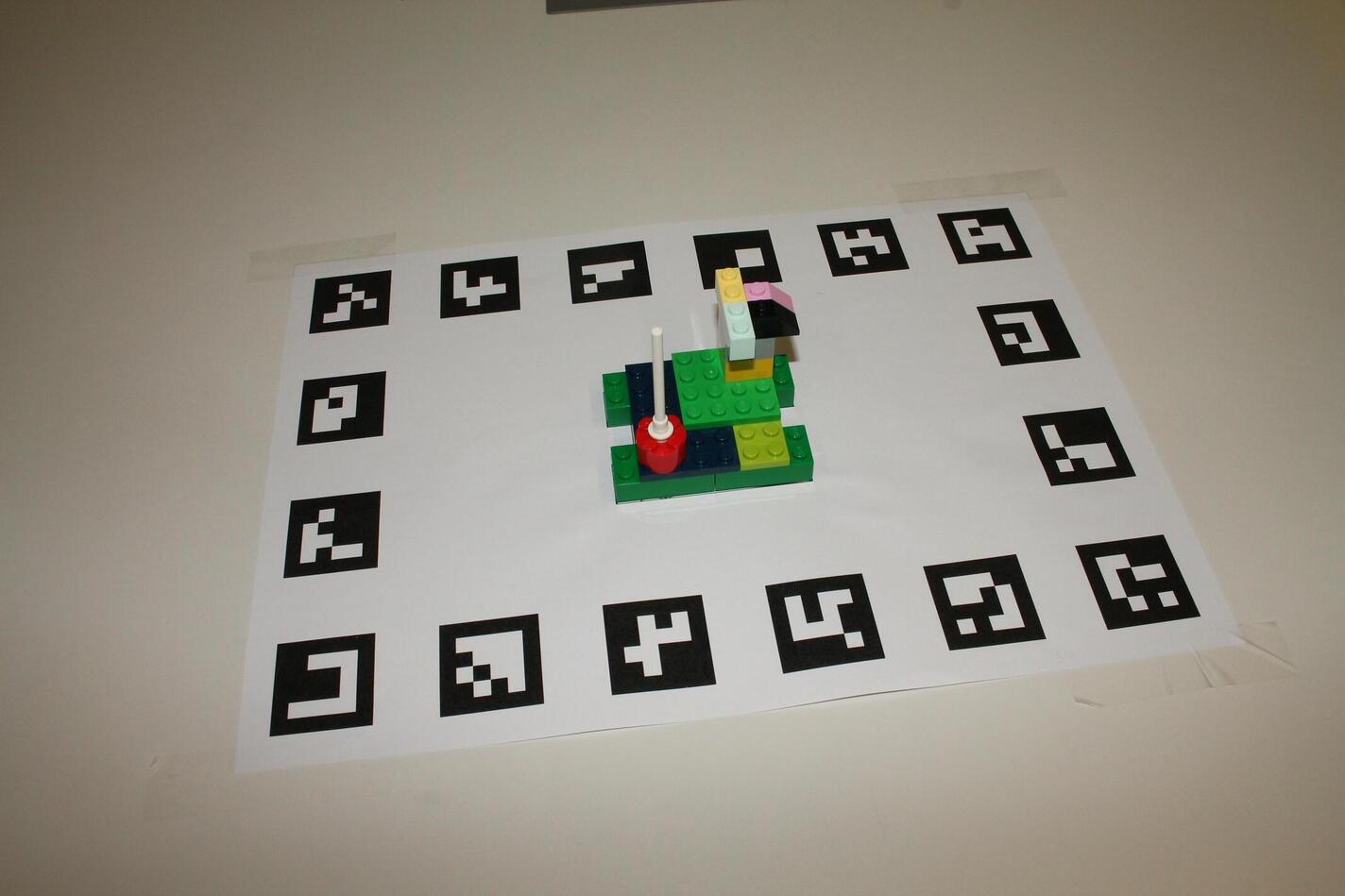}
        \centerzoom[0.25]{0.30\textwidth}{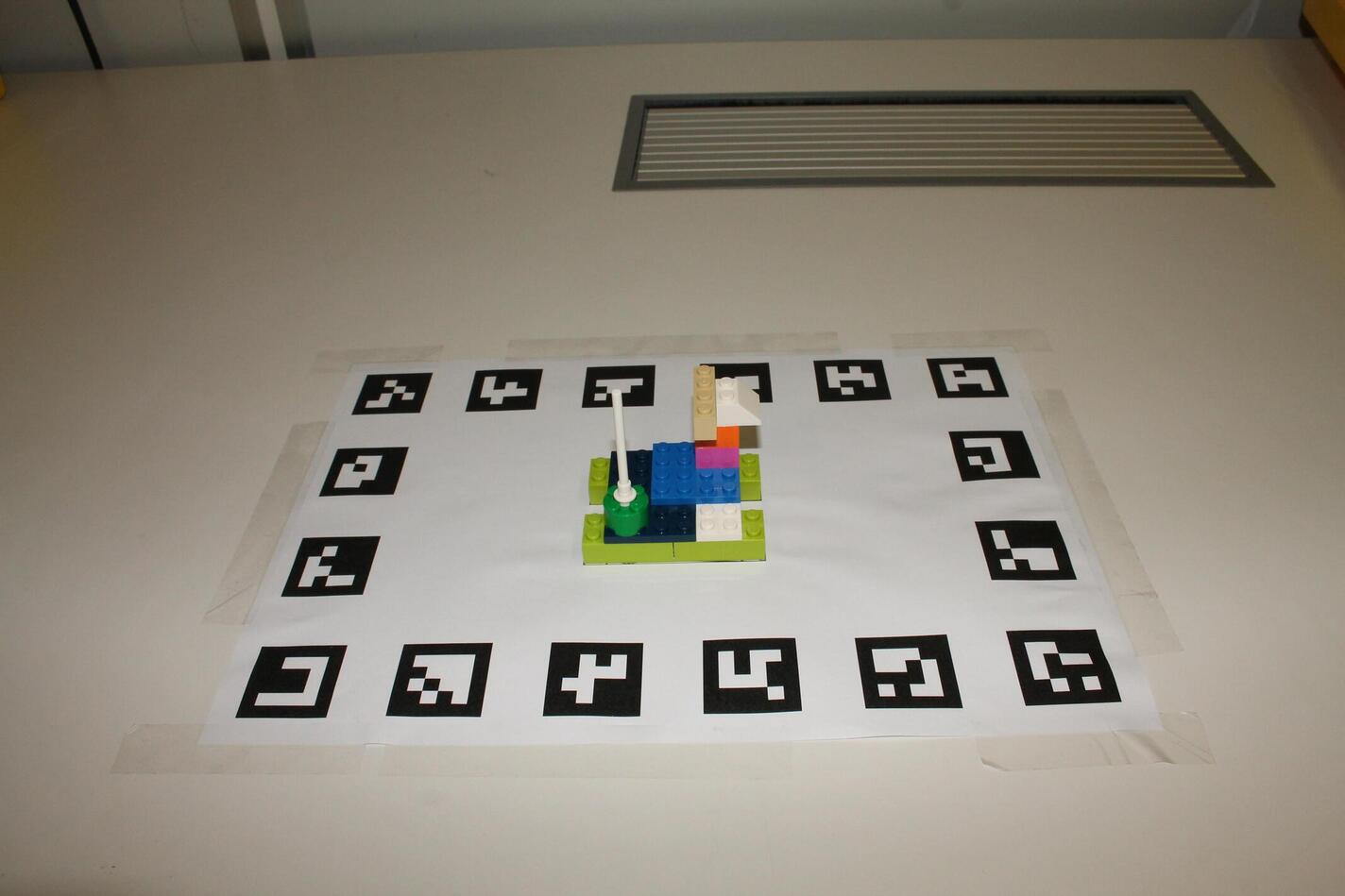}\\[0.2em]
        \centerzoom[0.15]{0.30\textwidth}{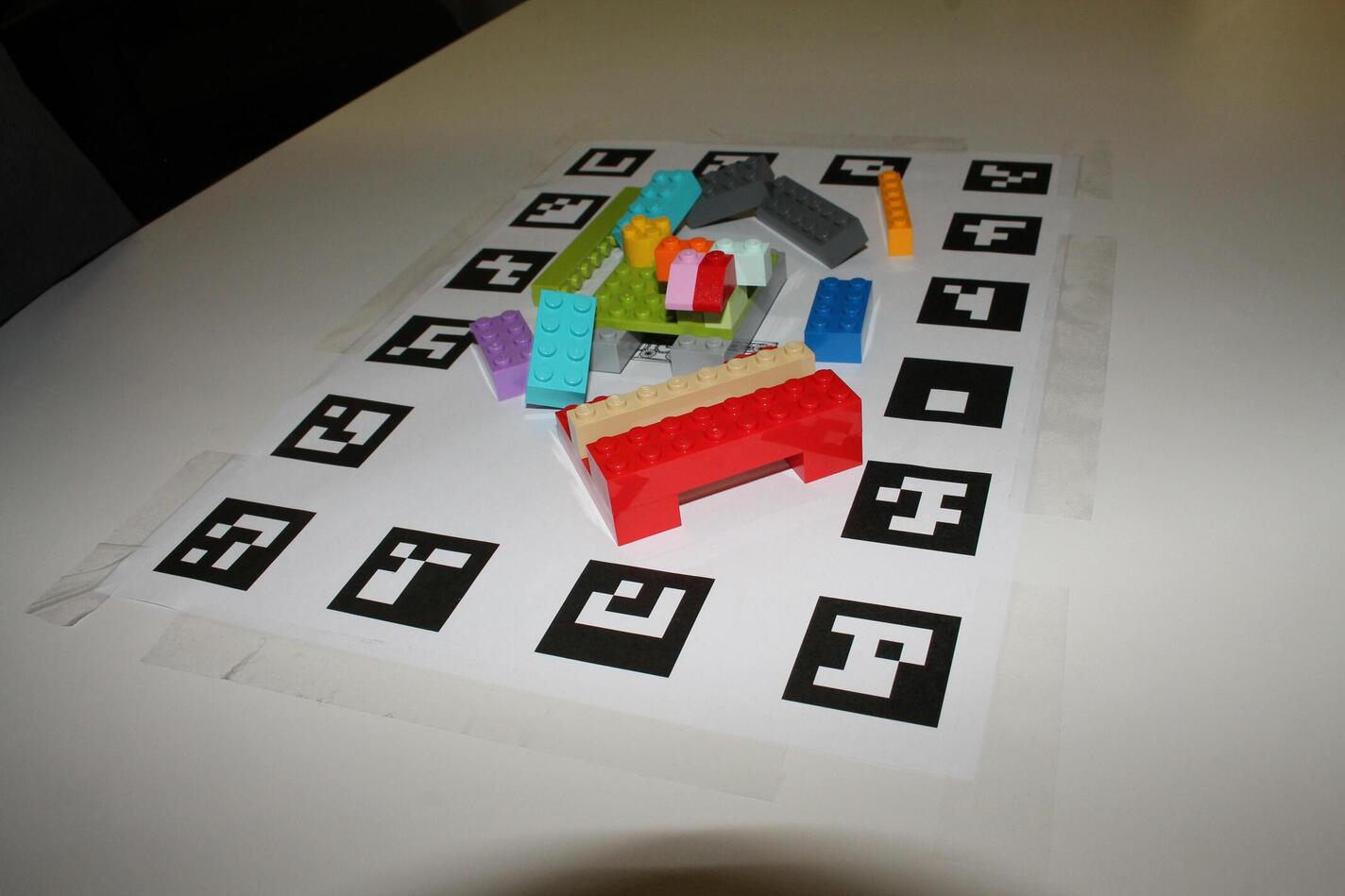}
        \centerzoom[0.15]{0.30\textwidth}{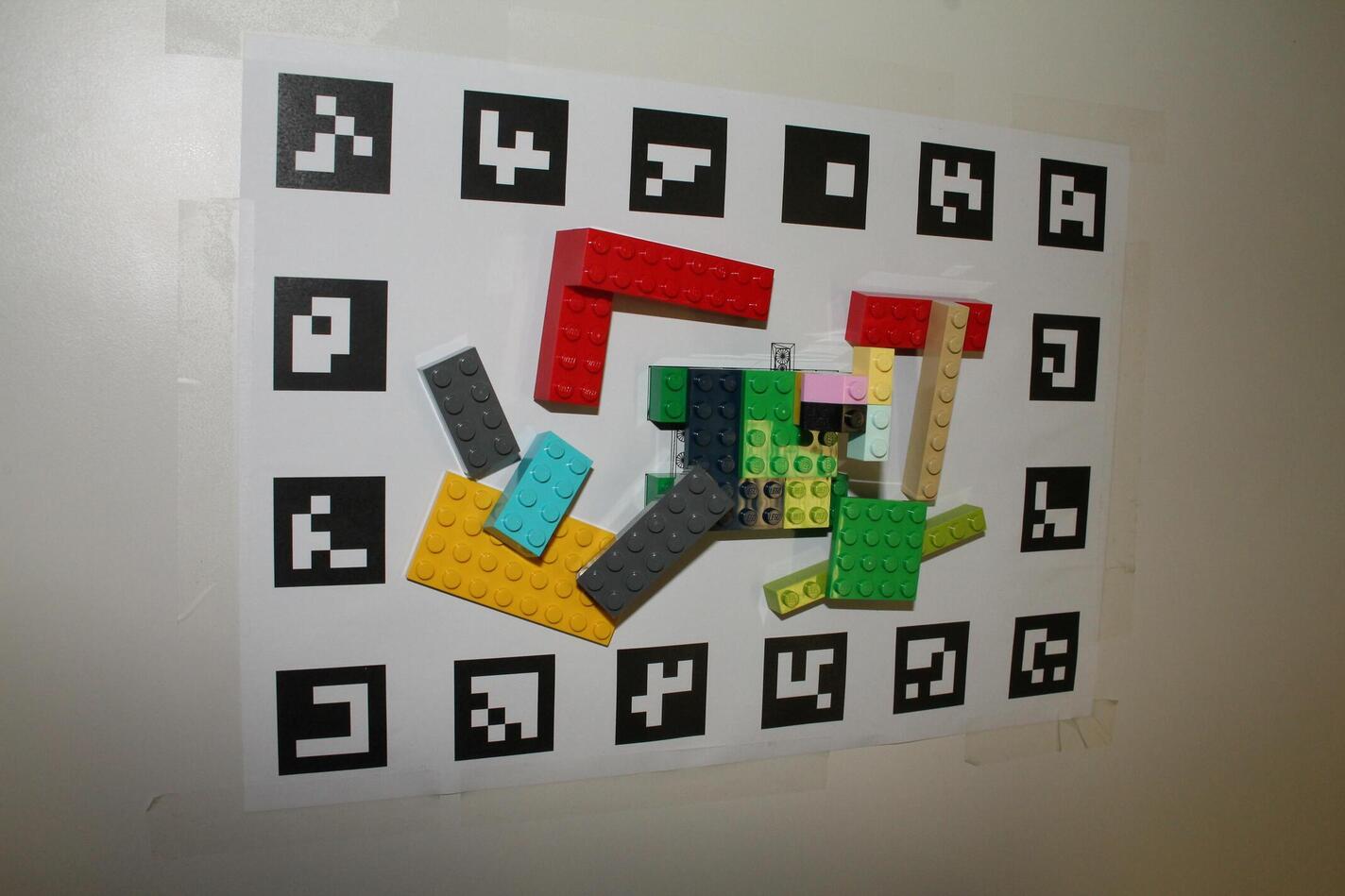}
        \centerzoom[0.15]{0.30\textwidth}{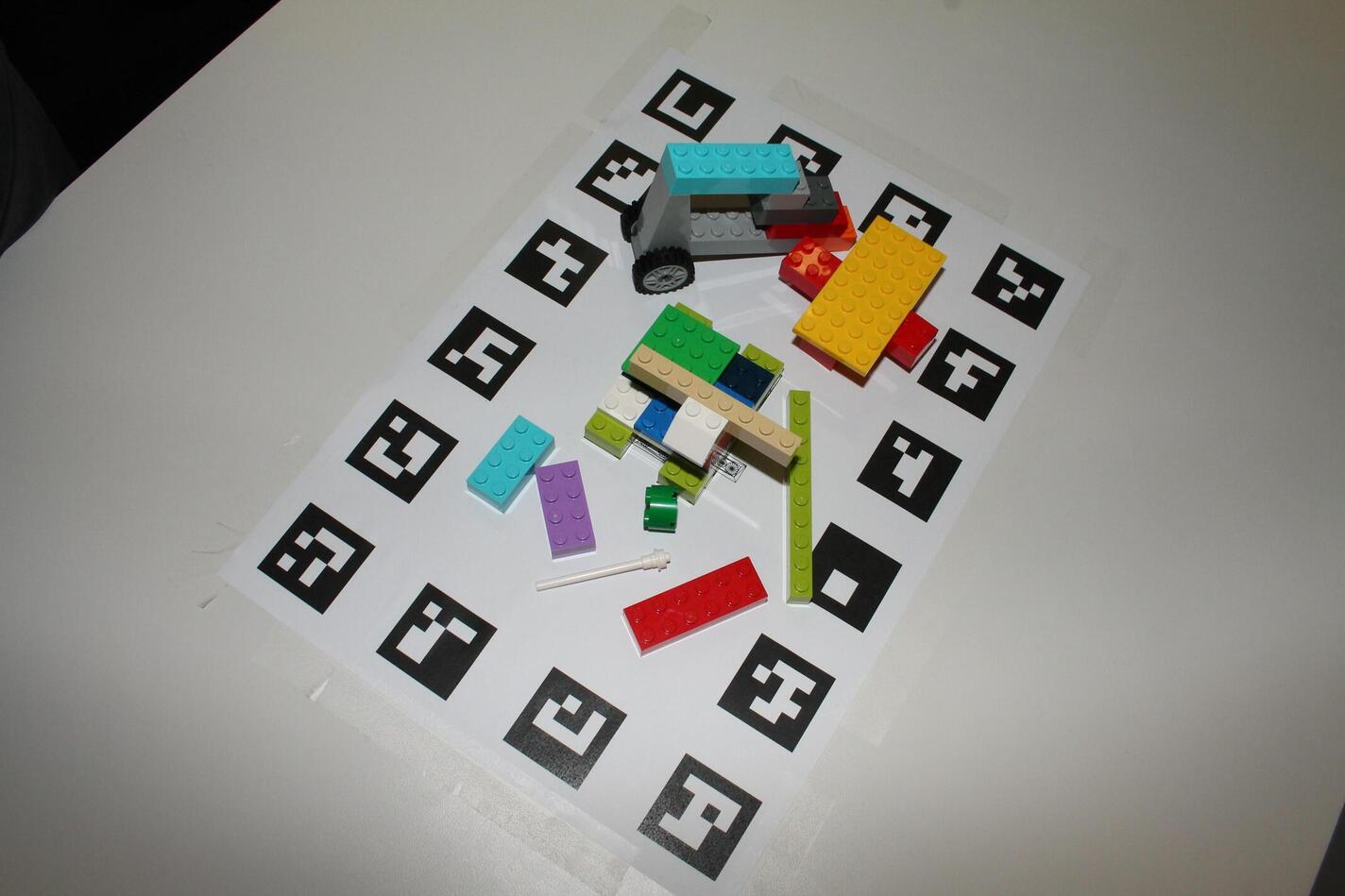}\\[0.2em]
        \caption{The top row shows the same geometric model with different textures, 
        while the bottom row shows variations of the model with assembly deviations, cluttered backgrounds, and different textures.
        }
        \label{fig:custom_variations}
    \end{subfigure}\\[0.9em]

    \begin{subfigure}{\linewidth}
        \centering
        \includegraphics[width=0.48\textwidth]{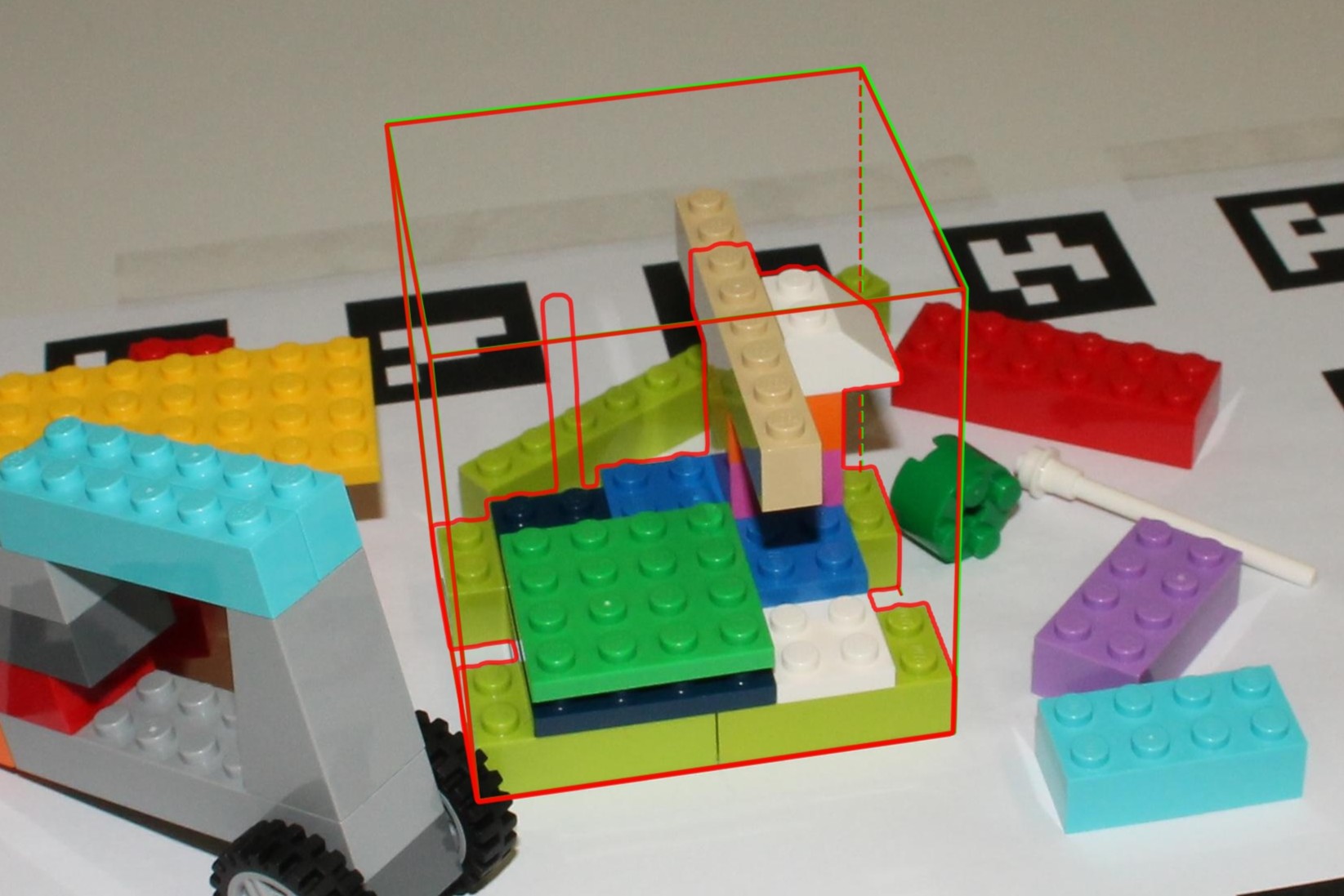}
        \includegraphics[width=0.48\textwidth]{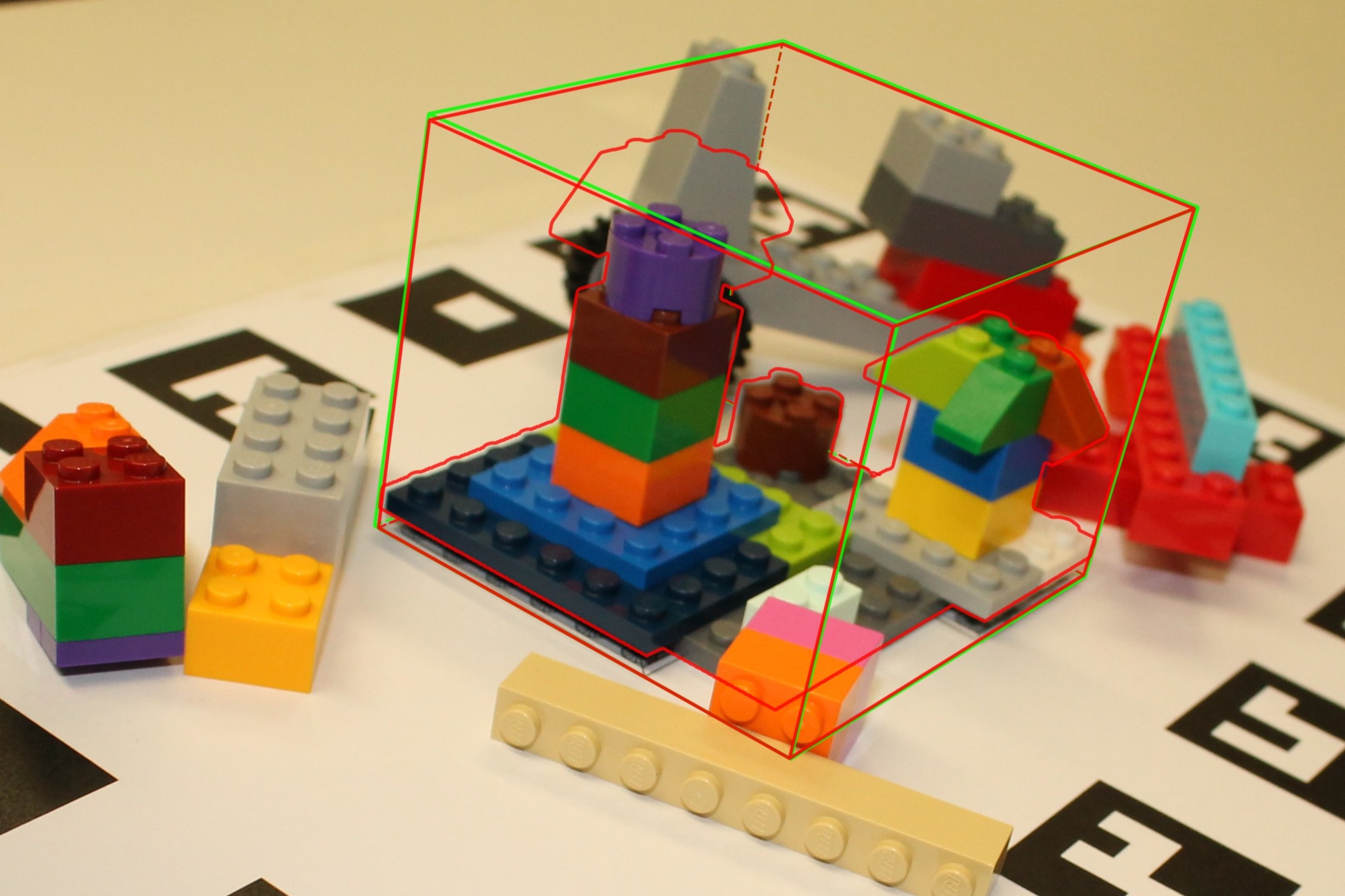}\\[0.2em]
        \includegraphics[width=0.48\textwidth]{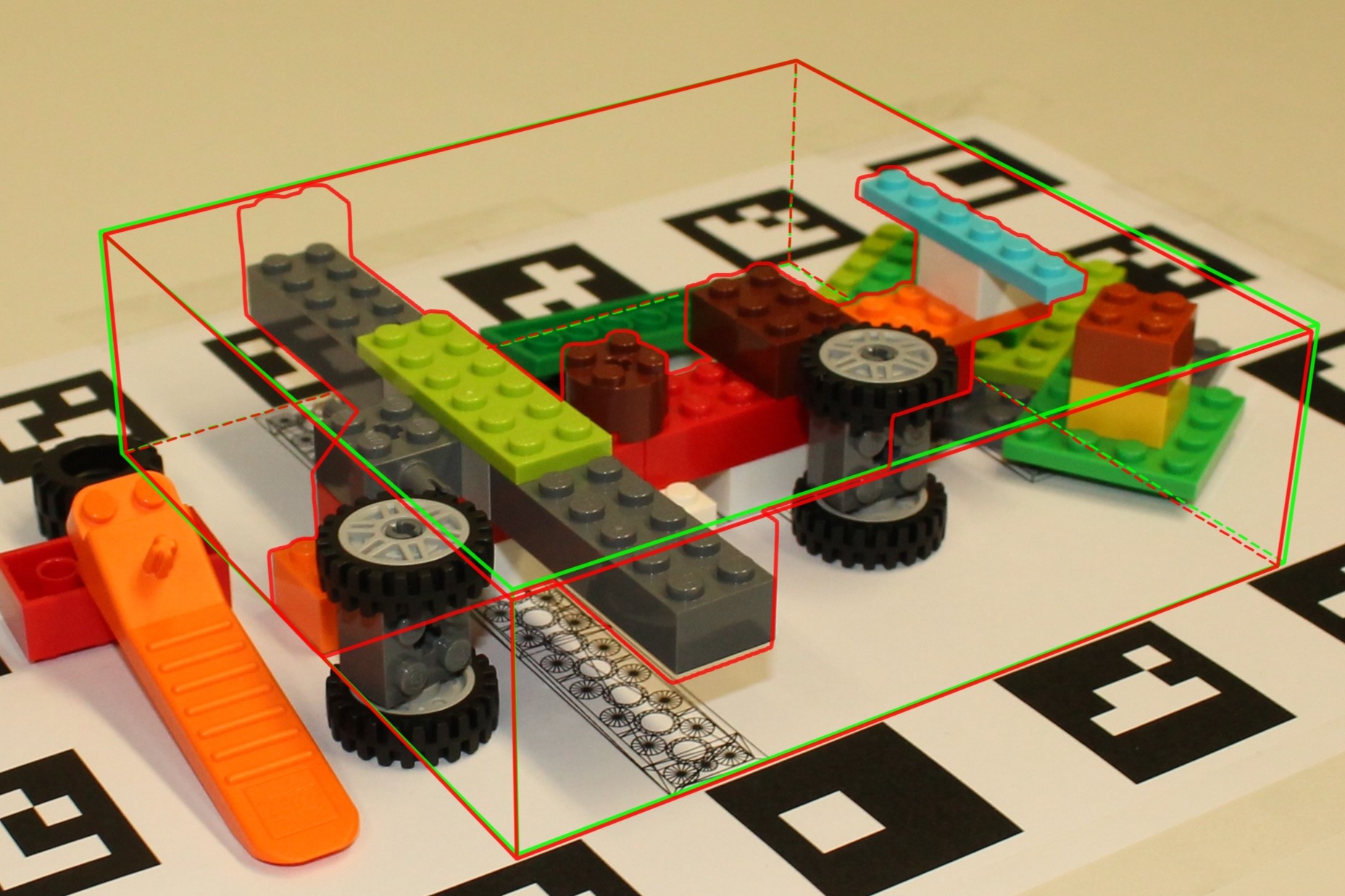}
        \includegraphics[width=0.48\textwidth]{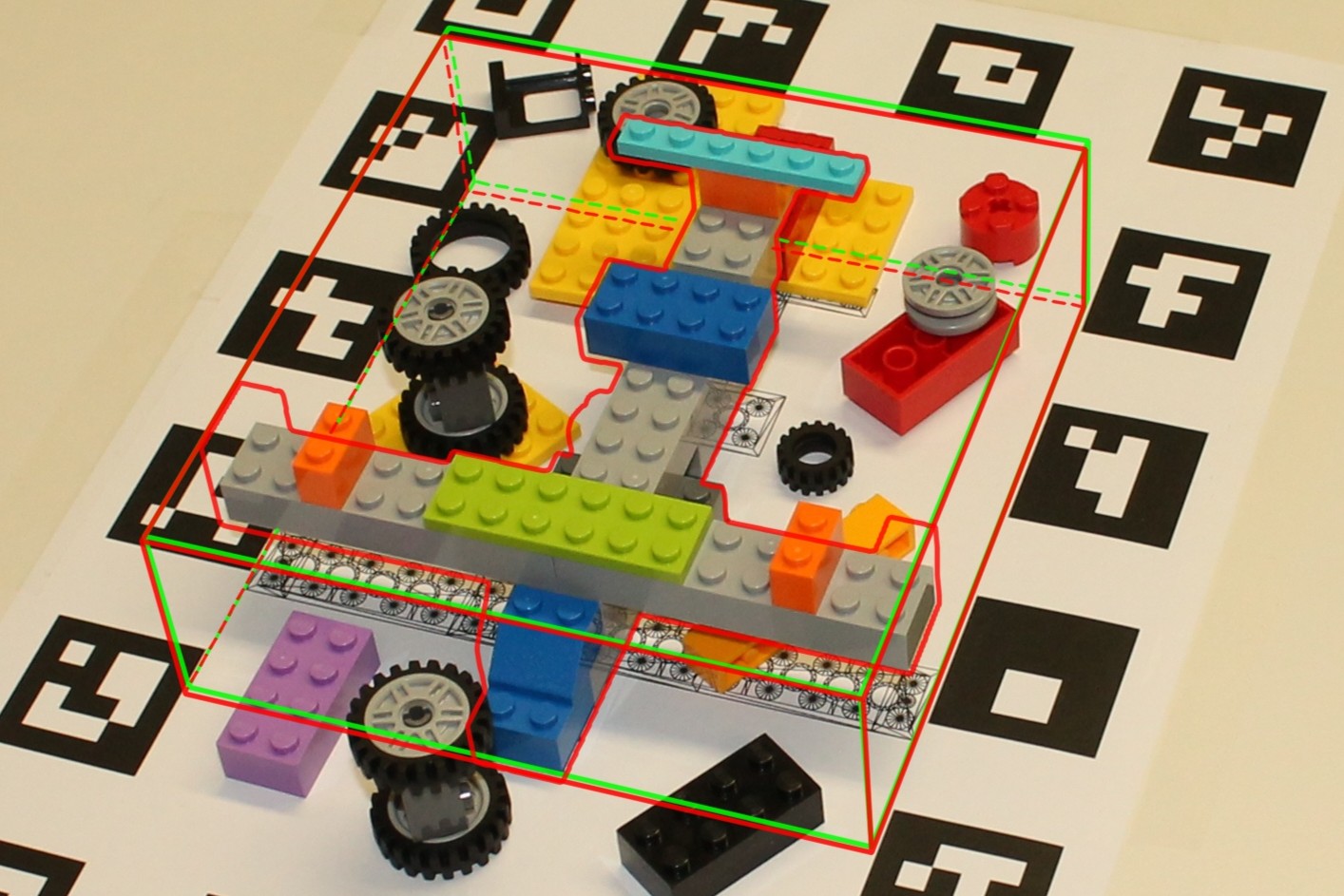}
        \caption{Results of our \textbf{PIXIE} method on our custom dataset, showing the estimated pose of the \textit{target model} (rendered in red) and the ground truth pose (rendered in green).
        All results are obtained using the same single reference set per object, generated from the \textit{target model} without defects or knowledge about the model textures.
        }
        \label{fig:custom_internet}
    \end{subfigure}

    \caption{Examples from our custom dataset: texture variations (b) and assembly deviations (a, b).
    PIXIE successfully estimates poses across all conditions using only the \textit{target model}, without knowledge of textures or defects (c).
    }
    \label{fig:custom_dataset_results}
\end{figure}

We used LEGO bricks as a controlled proxy: repetitive structures make geometric matching challenging, while enabling systematic texture and assembly-variation generation.
One scene describes a specific assembly configuration of a given object, and multiple images are captured from different viewpoints (Figure~\ref{fig:custom_dataset_results}).

Ground truth poses are obtained via marker-based calibration (1.6\,pixel average reprojection error). Calibrated intrinsics and ground truth \textit{defective models} are also provided, leading to a total of 48 different object models.
For each scene, our method is provided with only the untextured \textit{target model} and must estimate the pose of the object in the query image. 
The object instance in the query image may correspond to the \textit{target model} (plus texture and color variations) or one of the \textit{defective models} (geometric deviations and texture variations).
This setup tests a method's ability to perform pose estimation under both ideal conditions and conditions with geometric deviations, without any model-specific training or prior knowledge of the textures or types of defects present.
All methods compared in this section use the provided bounding boxes to crop reference and query images, ensuring a fair comparison.
To quantify assembly defects and geometric deviations, we propose two complementary metrics:

\textbf{IoU3D:} Volumetric overlap between the \textit{target model} and \textit{defective models}, computed by voxelizing both and taking the intersection-over-union ratio~\cite{ravi2020pytorch3d}. 

\textbf{Average Perspective Similarity (APS):} Measures the \emph{visual} impact of geometric deviations from a given camera pose $P$.
It extends the Visible Surface Discrepancy (VSD) metric by jointly accounting for both the pixel-wise overlap and the per-pixel depth differences between rendered views.
Specifically, depth maps are first compared pixel-wise over the mutually visible region $\mathcal{V}$, and the mean per-pixel similarity is then scaled by the silhouette IoU to penalise non-overlapping regions.
This two-factor design is particularly relevant when the silhouettes of the \textit{target model} and \textit{defective model} differ, as pure per-pixel averaging over the intersection would ignore the extent of non-overlapping regions.

Let $\mathcal{V} = \{(u,v) \mid D_i(u,v)>0 \wedge D_d(u,v)>0\}$ be the set of pixels visible in both renders, and let $\langle\cdot\rangle_{\mathcal{V}}$ denote the mean over $\mathcal{V}$.
Depth maps of the \textit{target model} and \textit{defective model} are rendered and compared pixel-wise:
\begin{align}
D_i &= \operatorname{depth\_render}(M_{\text{target}},\; P), \nonumber\\
D_d &= \operatorname{depth\_render}(M_{\text{defective}},\; P_{\text{gt}}), \nonumber\\
\delta(u,v) &= \min\!\left(\tfrac{|D_i(u,v) - D_d(u,v)|}{D_d(u,v)},\; 1\right), \nonumber\\
\operatorname{APS}(P) &= \operatorname{IoU}(D_i, D_d) \cdot \bigl\langle 1 - \delta \bigr\rangle_{\mathcal{V}}
\label{eq:aps}
\end{align}
Per-pixel depth differences are clipped at 100\%, pixels visible in only one render contribute to the IoU denominator but not to the mean in $\mathcal{V}$.
Setting $P{=}P_{\text{gt}}$ isolates geometric deviation independently of pose accuracy and is used to characterize the dataset (Table~\ref{tab:dataset_summary}).
Setting $P{=}P_{\text{pred}}$ additionally penalizes pose errors and serves as a pose evaluation metric.

\begin{table}[t]
\centering
\small
\begin{tabular}{lcccccc}
\hline
\#M & \#Sc. & \#Img. & \#T. & \#D. & APS (dev) & IoU3D (dev) \\
\hline
1 & 12 & 148 & 3 & 6 & 0.98\,(0.96) & 0.94\,(0.84) \\
2 & 12 & 169 & 3 & 7 & 0.92\,(0.85) & 0.87\,(0.77) \\
3 & 12 & 187 & 3 & 6 & 0.97\,(0.94) & 0.94\,(0.89) \\
4 & 12 & 191 & 3 & 6 & 0.96\,(0.92) & 0.95\,(0.89) \\
\hline
\textbf{$\Sigma$} & \textbf{48} & \textbf{695} & \textbf{12} & \textbf{25} & 0.96\,(0.91) & 0.93\,(0.85) \\
\hline
\end{tabular}
\caption{Custom Dataset Summary. Values in parentheses indicate the metric considering only models with geometric deviations.
Left to right: \#M = model index, \#Sc. = number of scenes, \#Img. = number of images, \#T. = number of textured instances, \#D. = number of assemblies with geometric deviations (defects).}

\label{tab:dataset_summary}
\end{table}

\begin{table*}[t]
\vspace{4pt}
\centering
\begin{threeparttable}
\begin{tabular}{l l c c | c c c c c | c}
    \toprule
    \textbf{\#} & \textbf{Method} & \textbf{3D model} & \textbf{\# ref. views} & \textbf{avg ADD (cm)} & \textbf{med ADD (cm)} & \textbf{ADD$_{0.5cm}$} & \textbf{ADD$_{1cm}$} & \textbf{ADD$_{5cm}$} & \textbf{APS} \\

\midrule
\textbf{1} & \textbf{PIXIE (ours)} & \textbf{target} & 121 & \cellcolor{green!20}\textbf{1.1172}  & \cellcolor{green!20}\textbf{0.3578} & \cellcolor{green!20}\textbf{61.01\%} & \cellcolor{green!20}\textbf{73.53\%} & \cellcolor{green!20}\textbf{92.95\%} & \cellcolor{green!20}\textbf{0.8674}\\
2 & PIXIE - depth only & target & 121 & 1.2429  & 0.3879 & 52.66\% & 66.33\% & 85.18\% & 0.8528 \\
3 & PIXIE - normal only & target & 121 & 1.1021 & 0.3628 & 55.68\% & 67.77\% & 86.47\% & 0.8596 \\
4 & GigaPose\tnote{2}~\cite{nguyen2024gigapose} & target & 161 & 15.0572 & 5.6063 & 0.29\% & 3.45\% & 31.37\% & 0.4622 \\
5 & FoundPose\tnote{2}~\cite{ornek2024foundpose} & target & 798 & 3.3884 & 1.6857 & 12.09\% & 31.08\% & 81.87\% & 0.7149 \\
\midrule
\textbf{6} & \textbf{PIXIE (ours)} & \textbf{defective} & 121 & \cellcolor{green!20}\textbf{1.0590} & \cellcolor{green!20}\textbf{0.2674} & \cellcolor{green!20}\textbf{66.86\%} & \cellcolor{green!20}\textbf{82.70\%} & \cellcolor{green!20}\textbf{95.93\%} & \cellcolor{green!20}\textbf{0.8659} \\
7 & GigaPose\tnote{2}~\cite{nguyen2024gigapose} & defective & 161 & 15.1194 & 5.7738 & 0.43\% & 3.02\% & 32.52\% & 0.4551 \\
8 & FoundPose\tnote{2}~\cite{ornek2024foundpose} & defective & 798 & 3.5282 & 1.5558 & 14.39\% & 35.11\% & 83.45\% & 0.7391 \\
\bottomrule
\end{tabular}
\begin{tablenotes}\footnotesize
    \item[2]{Using the official implementation with default settings.}    
\end{tablenotes}
\caption{Comparison on our custom dataset. Rows 1--5 use the \textit{target model} (potentially deviating from the physical object due to assembly defects) to evaluate pose estimation performance under ideal conditions. 
Rows 6--8 use \textit{defective models} (the models corresponding to the actual physical objects with assembly defects) to isolate the impact of geometric deviations on pose estimation performance.
All methods generate their own reference views offline.} 

\label{table:custom_dataset_results}
\end{threeparttable}
\end{table*}

PIXIE achieved median Average Distance of Model Points (ADD)~\cite{hinterstoisser2012model} of 0.36\,cm when using the \textit{target model} (row 1), improving to 0.27\,cm (row 6) when using the \textit{defective models}. 
This confirmed that geometric deviations measurably impacted accuracy, but the geometry-only approach remained robust even without knowledge of defects.
Compared to GigaPose~\cite{nguyen2024gigapose} and FoundPose~\cite{ornek2024foundpose}, our method significantly outperformed both in terms of ADD and APS, as well as Average Recall as proposed by the BOP benchmark~\cite{hodan2018bop}, 
demonstrating the strength of geometry-only pose estimation in this challenging setting with significant texture variations and geometric deviations (Table~\ref{table:custom_dataset_results}).
These results are achieved with significantly fewer reference views (121 vs. 161 for GigaPose and 798 for FoundPose), which is a direct consequence of our method's ability to leverage geometric cues effectively.

\subsection{Failure Cases and Limitations}
\label{subsec:failure_cases}

While our method demonstrates strong performance across various scenarios, several fundamental limitations exist 
due to the reliance on geometric distinctiveness.

\textbf{Geometric ambiguity:}
Simple or highly symmetric shapes (e.g., spheres, cylinders) yield few distinctive geometric features, 
producing ambiguous or unstable correspondences and multiple equally valid poses that geometry alone cannot distinguish 
(cf. two cylinders, top right of Fig. \ref{fig:bop_results}).
In our tests, we observed that in particular objects with little geometric features (e.g. the ape, screwdriver objects from the LineMOD Occluded dataset) are more difficult
to align, and the method can fail due to a lack of reliable correspondences.
The combination of low-resolution images, textureless objects, and heavy occlusion results in the LM-O dataset provide insufficient geometric 
detail for reliable matching, leading to lower performance compared to appearance-based methods that can leverage texture cues.

\textbf{Model-to-real geometric deviations:}
Assembly defects, manufacturing tolerances, wear, and damage may introduce geometric discrepancies, violating the correspondence assumptions of the method.
While we have shown that our method is robust to a certain degree of geometric deviations, especially for feature-poor objects, large deviations may prevent reliable correspondences and lead to inaccurate pose estimates.

\textbf{Computational cost:}
The current pipeline requires approximately 3.6\,s per image on an NVIDIA RTX 4090 GPU with an AMD Ryzen 9 7950X CPU and 64\,GB of RAM, primarily due to cross-modality matching across multiple viewpoints and in-plane rotations.
In our target industrial workflow, accuracy and robustness are prioritized over latency. Exploiting coarse pose priors can substantially reduce runtime.

\section{Conclusion}
\label{sec:conclusion}
This work presents a zero-shot framework for 6D pose estimation that operates exclusively on geometric 
information, requiring only untextured CAD models while eliminating the need for object-specific training, 
textured models, or explicit knowledge about assembly defects and geometric deviations.
By leveraging cross-modality feature matching between geometric renderings and query images, 
the method achieved state-of-the-art results on textureless objects in the BOP benchmark~\cite{hodan2018bop},
and demonstrated robustness to geometric deviations and occlusions on a novel dataset.
These results highlight the potential of geometry-only approaches for 6D pose estimation in challenging real-world scenarios, 
particularly in industrial settings where objects change frequently, textured models are unavailable, 
and assembly defects introduce geometric deviations from the 3D model.

To our knowledge, PIXIE is the first training-free, geometry-only pipeline for zero-shot 6D pose estimation from RGB images
that is robust to both texture and geometric deviations, and does not require any object-specific training.

Future work could explore how moving away from foundational cross-modality feature matchers and towards more 
specialized geometric matching architectures could further improve performance, especially in scenarios with significant
geometric deviations or occlusions. Additionally, integrating geometric and appearance cues could yield a more general-purpose 
pose estimator that maintains robustness in texture-less scenarios while leveraging texture when available.


{\small
\bibliographystyle{ieeetr}
\bibliography{main}
}


\end{document}